\documentclass[runningheads,a4paper]{llncs}

\usepackage{amssymb}
\usepackage{amsmath}
\usepackage{graphicx}
\usepackage{hyperref}
\usepackage{makecell}
\usepackage{xcolor}
\usepackage{url}
\usepackage[absolute,overlay]{textpos}

\begin{document}

\mainmatter  

 \begin{textblock*}{20cm}(0.5cm,0.5cm)
 \textcolor{red}{\noindent This is a preprint of the following chapter: 
 \\ X. Zhao, Probabilistic Robustness in Deep Learning:
 A Concise yet Comprehensive Guide,
 \\published in the book \textit{Adversarial Example Detection and Mitigation Using Machine Learning}, 
 \\edited by Ehsan Nowroozi, Rahim Taheri, Lucas Cordeiro, 2025, Springer Nature. 
 \\The final authenticated version will available online soon}
\end{textblock*}

\title{Probabilistic Robustness in Deep Learning: \\ A Concise yet Comprehensive Guide}

\titlerunning{Probabilistic Robustness in Deep Learning}

%
%
\author{Xingyu Zhao%
\thanks{This work is supported by the UK EPSRC New Investigator Award through Harnessing Synthetic Data Fidelity for Assured Perception of Autonomous Vehicles and the NVIDIA Academic Grant Program.}%
}
\authorrunning{X.~Zhao}

\institute{WMG, University of Warwick\\
Coventry, CV4 7AL, United Kingdom\\
\email{xingyu.zhao@warwick.ac.uk}
}

%
%

\toctitle{Lecture Notes in Computer Science}
\tocauthor{Authors' Instructions}
\maketitle

\begin{abstract}
Deep learning (DL) has demonstrated significant potential across various safety-critical applications, yet ensuring its robustness remains a key challenge. While adversarial robustness has been extensively studied in worst-case scenarios, probabilistic robustness (PR) offers a more practical perspective by quantifying the likelihood of failures under stochastic perturbations. This paper provides a concise yet comprehensive overview of PR, covering its formal definitions, evaluation and enhancement methods. We introduce a reformulated ``min-max'' optimisation framework for adversarial training specifically designed to improve PR. Furthermore, we explore the integration of PR verification evidence into system-level safety assurance, addressing challenges in translating DL model-level robustness to system-level claims. Finally, we highlight open research questions, including benchmarking PR evaluation methods, extending PR to generative AI tasks, and developing rigorous methodologies and case studies for system-level integration.
\end{abstract}

\section{Introduction}
\label{sec_intro}

Deep learning (DL) has demonstrated transformative potential across various industries, including transportation, healthcare, and energy. Its ability to process complex data and make accurate predictions makes it a promising candidate for integration into safety-critical applications \cite{tambon2022certify,rabe2021development,rech2024artificial}, such as autonomous driving, medical diagnosis, and energy grid management. However, the \textit{robustness} of DL models in these high-stakes environments remains a significant concern, posing a risk to their safe deployment \cite{huang2017safety,bloomfield2019disruptive,zhao2020safety}.

Robustness in DL generally refers to a model's ability to maintain consistent predictions despite small input perturbations \cite{carlini2017towards,huang_survey_2020,dong2023reliability}. In other words, a robust model should produce the same output even when its input is slightly altered. Adversarial examples (AEs) \cite{szegedy2013intriguing,goodfellow2014explaining} are inputs that have been perturbed to cause a model to make incorrect predictions. These perturbations are often imperceptible to humans but can lead to significant errors in the DL model's output. Indeed, robustness has become a central focus in DL research, encompassing a wide range of topics like the study of adversarial attacks in malicious contexts \cite{goodfellow2014explaining,madry2018towards,carlini2017towards,zuhlke2024adversarial}, methods for detecting AEs by testing \cite{du2019deepstellar,huang2023hierarchical}, and techniques for verifying model robustness with formal guarantees \cite{huang2017safety,ruan2018reachability}. Additionally, various approaches have been developed to improve robustness, such as distillation \cite{papernot2016distillation}, randomized smoothing \cite{cohen2019certified}, gradient masking \cite{athalye18a}, and adversarial training (AT) \cite{goodfellow2014explaining,madry2018towards}. As multi-modal foundation models gain popularity, recent research has increasingly shifted toward understanding their robustness \cite{du2024stable,zhang2024protip,zhang2024trustworthy}.

Notably, the majority of robustness research focuses on \textit{adversarial robustness} (AR), which addresses \textit{worst-case} and \textit{deterministic} scenarios. The terms ``worst-case'' and ``deterministic'' refers to the fact that AR metrics and their corresponding evaluation problems are typically formulated around the following three (distinct yet interrelated) types of questions \cite{zhang2024protip}: 
\begin{itemize}
    \item a binary question, ``Does an AE exist within the perturbation norm-ball?''
    \item an optimisation question on maximising the safe perturbation distance, ``What is the largest perturbation radius within which no AE exists?''
    \item an optimisation question on maximising the prediction loss, ``Which AE within the given norm-ball yields the highest prediction loss?''
\end{itemize}
In contrast, a more recent and \textit{practical} (cf.~Section \ref{sec_pr_def}) view considers \textit{probabilistic robustness} (PR) \cite{webb2018statistical,weng2019proven,wang2021statistically,pautov2022cc,couellan2021probabilistic,baluta2021scalable,tit2021efficient,zhang2022proa,robey2022probabilistically,pmlr_v206_tit23a,zhang2024protip}---it answers:
\begin{itemize}
    \item a statistical inference question, ``What is the \textit{probability} of discovering an AE when the input is stochastically perturbed according to a specified distribution within a given norm-ball?''
\end{itemize}
A ``frequentist'' interpretation of this probability is the \textit{limiting relative frequency} of perturbations for which the DL model's output remains unchanged, within an infinite sequence of independently generated perturbations \cite{zhang2024protip}. In other words, it represents the \textit{proportion} of non-AEs in the infinite set of all possible input perturbations. Thus, more intuitively, the statistical inference question of PR becomes ``What is the proportion of AEs in the given perturbation norm-ball?''

In this paper, our main contribution is to deliver a concise yet comprehensive overview of all critical aspects of PR research. This includes its formal definition and associated metrics, existing evaluation methods, improvement techniques, and its significance for system safety assurance and reliability modelling. Specifically, we begin by summarising and contrasting the key distinctions between PR and AR in Sec.~\ref{sec_pr_def}. Following this, we provide an in-depth analysis of existing evaluation methods for PR in Sec.~\ref{sec_pr_eva}. In Sec.~\ref{sec_pr_enhance}, we revisit existing AT techniques and propose a reformulated ``min-max'' optimisation framework as the theoretical foundation for AT specifically designed to enhance PR. Furthermore, from a systems engineering perspective, we explore potential pathways for integrating PR verification evidence into AI system-level assurance to enhance safety and reliability in Sec.~\ref{sec_assurance}. Finally, we identify open questions and key challenges in the field, outlining promising directions for future research in Sec.~\ref{sec_challenges}.

\section{Definition of Probabilistic Robustness}
\label{sec_pr_def}

Although definitions of robustness vary across different DL tasks and model types, it generally refers to a DL model’s ability to maintain consistent predictions despite small input perturbations. Typically it is defined as all inputs in a region $\eta$ have the same prediction, where $\eta$ is a small norm ball (in a $L_p$-norm distance) of radius $\gamma$ around an input $x$. A perturbed input (e.g., by adding noise on $x$) $x'$ within $\eta$ is an AE if its prediction label differs from~$x$.

\begin{figure}[t!]
\centering
\includegraphics[width=0.78\linewidth]{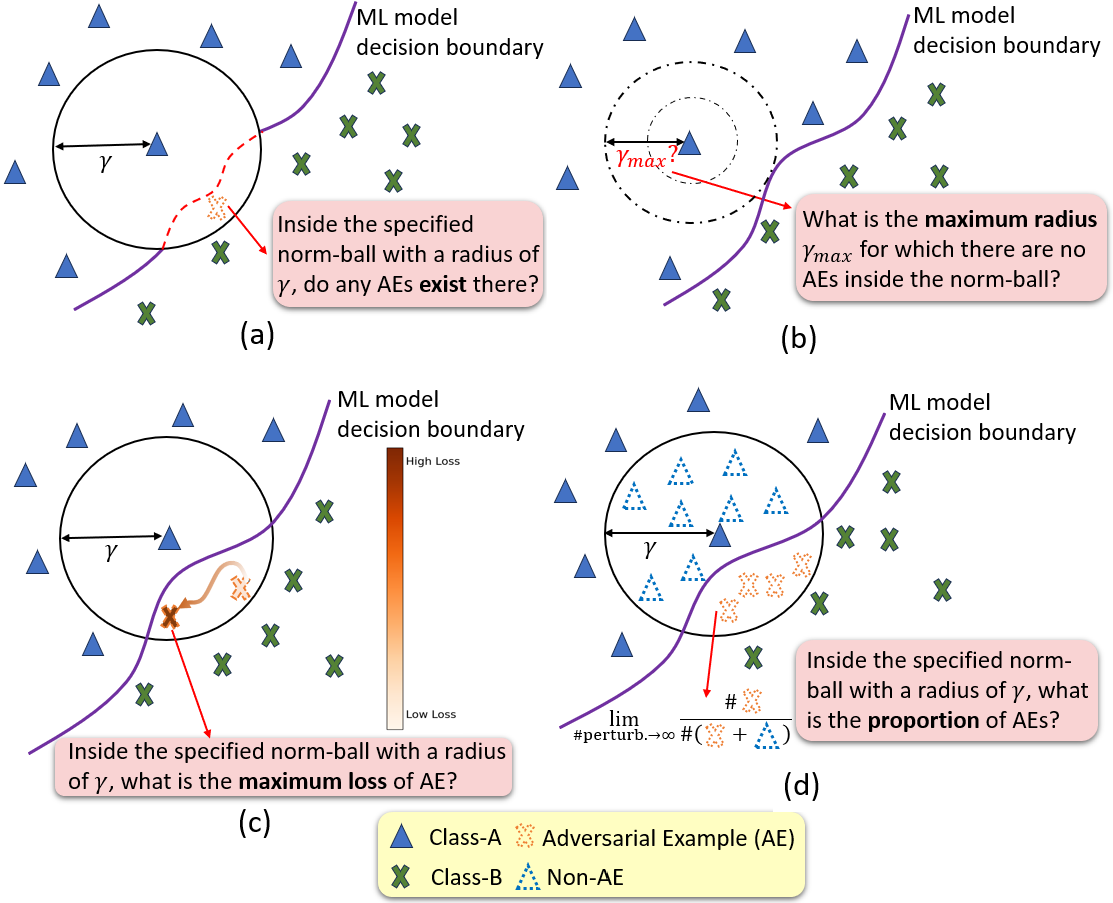}
\caption{Four types of problem formulations of robustness evaluation problems \cite{zhang2024protip}.}
\label{fig_wcr_pr}
\end{figure}

Robustness are typically measured on metrics defined around AEs \cite{huang_survey_2020,zuhlke2024adversarial}. As discussed earlier in Sec.~\ref{sec_intro}, four types of robustness metrics, as summarised in \cite{zhang2024protip,dong2023reliability} and Fig.~\ref{fig_wcr_pr}, address: \textit{1)} a binary question, e.g., formulated as a satisfiability problem that usually can be solved by SAT/SMT solvers \cite{huang2017safety,katz2017reluplex}; \textit{2)} a maximum safe perturbation radius question, e.g., formulated as reachability analysis problems \cite{zhang2018efficient,ruan2018reachability}; \textit{3)} a maximum prediction loss question, e.g., formulated as adversarial attacks leveraging gradient information \cite{goodfellow2014explaining,madry2018towards,carlini2017towards,croce2020reliable}; and \textit{4)} a statistical inference question estimating the ``proportion'' (a \textit{limiting relative frequency} interpretation of the probability of detecting AEs when generating random perturbations) of AEs within a given perturbation norm-ball.
\begin{definition}[Probabilistic Robustness]
\label{def_Prob_rob_classification}
For a DL model $f_\theta$ that takes input $x$, the local PR of an input $x$ in a norm ball of radius $\gamma$ is:
\begin{align}
\label{eq_probabilistic_robustness_definition}
\textit{PR} (x, \gamma) = & \mathbb{E}_{\substack{x' \sim Pr(\cdot \mid x)\\x':\|x-x'\|\leq \gamma}}[I_{\{f_\theta(x')=f_\theta(x)\}}(x')] 
\end{align}
where $I_{\mathcal{S}}(x)$ is an indicator function---it is equal to 1 when $\mathcal{S}$ is true and 0 otherwise; $Pr(\cdot \mid x)$ is the local distribution of how perturbations $x'$ are generated from $x$, which is precisely the ``input model'' used by, e.g., \cite{webb2018statistical,weng2019proven,zhang2024protip}.
\end{definition}
While PR in Eq.~\eqref{eq_probabilistic_robustness_definition} represents the \textit{expected probability} of detecting AEs, alternative tail-sensitive local PR measures can be defined, e.g., $\rho$-essential supremum ($\rho$-ess sup) for a given PR threshold and Conditional Value-at-Risk (CVaR) for tail averages \cite{robey2022probabilistically}, and Entropic Value-at-Risk (EVaR) for exponential tail bounds \cite{zhang2024prass}.

It is evident that cases (a), (b), and (c) in Fig.~\ref{fig_wcr_pr} focus on \textit{extreme scenarios}, where robustness is evaluated based on the existence of a \textit{deterministic} AE, reflecting a \textit{worst-case} perspective. This approach, originally termed ``adversarial" robustness, is particularly relevant in malicious contexts and \textit{security} studies. While a comprehensive review of AR is out of the scope of this paper, we refer readers to surveys \cite{huang_survey_2020,meng2022adversarial,zuhlke2024adversarial,chakraborty2021survey}. In contrast, case (d) adopts a \textit{probabilistic} view, assessing \textit{overall} local robustness by considering the presence of AEs statistically. Compared to AR, this probabilistic view of PR is more practical for real-world applications, as worst-case evaluations are often unnecessary, especially for large models operating in \textit{benign} environments. Quantifying the proportion/risk of AEs is more relevant for \textit{reliability and safety}, where ensuring the AE risk remains below an acceptable threshold is sufficient, rather than requiring it to be exactly zero \cite{webb2018statistical,wang2021statistically,zhang2022proa,zhang2024protip,robey2022probabilistically}.

Although some studies generate perturbations \textit{stochastically}, this does not imply a focus on PR, e.g., randomised smoothing \cite{cohen2019certified} aim to certify a safe perturbation radius (i.e., case (b) in Fig.~\ref{fig_wcr_pr}) by replacing standard model inference with the averaging of predictions over a set of randomly perturbed inputs. 

Furthermore, the PR definition of Eq.~\ref{eq_probabilistic_robustness_definition} is a type of \textit{local} robustness focusing on model stability around an given \textit{individual input}, as opposed to \textit{global} robustness across the entire input space. A closely related concept to local PR but defined at the global level is the \textit{probabilistic Lipschitzness} \cite{mangal2019robustness,mangal2020probabilistic}: 
\begin{definition}[Probabilistic Lipschitzness]
\label{def_probabilistic_Lipschitzness}
Given a distribution $D$ over the entire input space of model $f$, its probabilistic Lipschitzness is formal stated as:
\begin{equation}
\label{eq_probabilistic_Lipschitzness}
Pr_{x,x'\sim D \bigotimes D}(||f(x)-f(x')||\leq k*||x-x'|| \mid ||x-x'||\leq \gamma ) \geq 1-\epsilon,
\end{equation}
which is defined globally by focusing on pairs of $\gamma$-close inputs such that their cumulative probability on the Lipschitz property is at least $1-\epsilon$.
\end{definition}

Similarly, \cite{wang2021statistically} explicitly defines global robustness probabilistically, referred to as \textit{total statistical robustness} (or simply termed ``reliability'' in \cite{zhao_assessing_2021,dong2023reliability}):
\begin{definition}[Total Statistical Robustness]
\label{def_total_sta_rob}
Reusing the notations from aforementioned definitions, the total statistical robustness of a model $f$ is:
\begin{equation}
\label{eq_total_sta_rob}
\mathbb{E}_{X\sim D}[\mathbb{E}_{X'\sim Pr(\cdot \mid x)}[I_{\{f(x')=f(x)\}}(x')]]
\end{equation}
which is the expectation of local PR over the distribution of model's input space.
\end{definition} 
Again, the ``inner-expectation'' can be replaced by other risk measures like $\rho$-ess sup and CVaR in \cite{robey2022probabilistically}, and EVaR in \cite{zhang2024prass}.


\section{Evaluation of Probabilistic Robustness}
\label{sec_pr_eva}

For AR, evaluation is typically conducted by measuring adversarial accuracy on a test set, where strong adversarial attacks are applied, such as those in \cite{croce2020robustbench,dong2020benchmarking}. While no benchmark for PR currently exists, we provide a summary of existing works that evaluate\footnote{We use the term ``evaluation'' broadly to encompass related concepts such as certification, verification, and assessment, without distinguishing their subtle differences. Here, ``evaluation'' refers to any study that evaluates a model's robustness based on PR metrics without modifying the DL model itself.} PR-related metrics in Definitions~\ref{def_Prob_rob_classification}, \ref{def_probabilistic_Lipschitzness}, and \ref{def_total_sta_rob} in Table~\ref{tab_eva}. 15 papers (including journal extension of 2 conference papers) are organised chronologically (since 2019) and explicitly categorised based on the specific PR metric studied, how input perturbations generated, the core techniques or theoretical foundations underlying their estimators, and the DL tasks considered.

\cite{webb2018statistical} introduces the first \textit{black-box} estimator based on adaptive multi-level splitting, a Monte Carlo technique for estimating the probability of rare events. Similarly \cite{tit2021efficient} proposes a more efficient black-box estimator by using a variant of adaptive multi-level splitting and a sequential Monte Carlo sampling method known as ``last particle". Another black-box estimator is presented in \cite{baluta2021scalable}, which combines Chernoff bounds with simple Monte Carlo sampling. 

In contrast, PROVEN \cite{weng2019proven} is the first \textit{white-box} analytical approach, leveraging the fact that a deep learning model’s output can be lower/upper-bounded by two linear functions, and then applying Hoeffding’s inequality to derive PR guarantees. Similarly, \cite{couellan2021probabilistic} presents a white-box method that derives upper bounds on tail probabilities using the Cramér-Chernoff concentration inequality to analyse uncertainty propagation through the DL model layers. More recently, \cite{pmlr_v206_tit23a} proposes a highly efficient estimator for local PR by integrating fast gradient computations with Hamiltonian Monte Carlo sampling.

While the aforementioned studies primarily focus on perturbations modeled as random \(L_p\) noise, \cite{zhang2022proa} and \cite{pautov2022cc} extend PR analysis to \textit{semantic perturbations} (referred to as functional perturbations in \cite{zhang2022proa}), e.g., rotation and translation. In \cite{zhang2022proa}, simple Monte Carlo sampling and an adaptive version of Hoeffding's inequality are employed to establish confidence bounds on local PR. In \cite{pautov2022cc}, a black-box approach is adopted, utilising the Cramér-Chernoff inequality alongside simple Monte Carlo sampling.

At the global-level, \cite{mangal2019robustness,mangal2020probabilistic} is the first work look at the property of Def.~\ref{def_probabilistic_Lipschitzness} which is checked probabilistically by an algorithm based on abstract interpretation and importance sampling techniques. The related but more straightforward metric of Definition~\ref{def_total_sta_rob}, was first introduced in \cite{wang2021statistically,zhao_assessing_2021}, where the latter refers to it as ``reliability''. In \cite{wang2021statistically}, the estimator from \cite{webb2018statistical} is used at the local level, with global estimation performed via simple Monte Carlo sampling (as its primary focus is on improving global robustness (cf. Section~\ref{sec_pr_enhance}) through data augmentation rather than estimation). In contrast, \cite{zhao_assessing_2021} explicitly connects this metric to the \textit{operational profile} (OP) concept from software reliability engineering \cite{musa1993operational}, approximating the OP from data after partitioning the input space.

\begin{table}[th]
\centering
\caption{Estimators on PR-related metrics: local PR (LPR, Def.~\ref{def_Prob_rob_classification}), global Probabilistic Lipschitzness (GPL, Def.~\ref{def_probabilistic_Lipschitzness}) and total statistical robustness (TSR, Def.~\ref{def_total_sta_rob}).}
\label{tab_eva}
\begin{tabular*}{1\linewidth}{c|c|c|c|c}
Paper & Metric & Perturb. & Estimator & DL Task \\ \hline \hline
ICLR'19 \cite{webb2018statistical}&LPR& $L_p$ noise &\makecell{adaptive multi-level splitting, \\Metropolis–Hasting algo.}&Img. classific.\\ \hline
ICML'19 \cite{weng2019proven}&LPR&$L_p$ noise&\makecell{layer-wise bound propagation,\\Hoeffding's inequality} &Img. classific.\\ \hline
ICSE'19 \cite{mangal2019robustness,mangal2020probabilistic} &GPL&$L_p$ noise&\makecell{abstract interpretation, \\ importance sampling}&Img. classific.\\ \hline
UAI'21 \cite{wang2021statistically}&TSR&$L_p$ noise&simple Monte Carlo \& \cite{webb2018statistical} &Img. classific.\\ \hline
AISafety'21 \cite{zhao_assessing_2021,dong2023reliability}&TSR&$L_p$ noise&\makecell{input space partition,\\ operational profile}&Img. classific.\\ \hline
Elsevier-NN \cite{couellan2021probabilistic}&LPR&$L_p$ noise&\makecell{Cramer-Chernoff inequality,\\layer-wise uncertainty propaga.}&Reg.\&classific.\\ \hline
ICSE'21 \cite{baluta2021scalable}&LPR&$L_p$ noise&\makecell{Chernoff bounds, \\ simple Monte Carlo}&Img. classific.\\ \hline
NeurIPS'21 \cite{tit2021efficient}&LPR&$L_p$ noise&\makecell{adaptive multi-level splitting,\\ ``last particle'' simulation}&Img. classific.\\ \hline
ECML'22 \cite{zhang2022proa}&LPR&semantic&\makecell{adaptive Hoeffding's inequality, \\ simple Monte Carlo}&Img. classific.\\ \hline
AAAI'22 \cite{pautov2022cc}&LPR&semantic&\makecell{Cramer-Chernoff inequality, \\simple Monte Carlo}&Img. classific.\\ \hline
ICML'22 \cite{robey2022probabilistically}&\makecell{$\rho$-ess sup\\CVaR}&$L_p$ noise&\makecell{stochastic gradient decent,\\simple Monte Carlo} &Img. classific.\\ \hline
AIStats'23 \cite{pmlr_v206_tit23a}&LPR&$L_p$ noise&\makecell{gradient-informed\\ Hamiltonian Monte Carlo}&Img. classific.\\ \hline
ICCV'23 \cite{Huang_2023_ICCV}&LPR&$L_p$ noise&\makecell{adaptive multi-level splitting, \\Metropolis–Hasting algo.}&XAI\\ \hline
IJCAI'24 \cite{zhang2024prass}&EVaR&$L_p$ noise&\makecell{adaptive stochastic
search,\\simple Monte Carlo}&Img. classific.\\ \hline
ECCV'24 \cite{zhang2024protip}&LPR&text char.&\makecell{adaptive Hoeffding's inequality, \\ sequential statistical testing}&Text2Image\\ \hline
\end{tabular*}
\end{table}

Most existing research on PR has focused on image classification tasks. The first and, to date, only extension of PR to explanations from eXplainable AI (XAI) methods is \cite{Huang_2023_ICCV}. In \cite{Huang_2023_ICCV}, the PR of Def.~\ref{def_Prob_rob_classification} is redefined in terms of wrong explanations after random perturbations of images, it applies the same techniques as \cite{webb2018statistical}. More recently, the first study on PR in text-to-image generative AI was introduced in \cite{zhang2024protip}. This work employs the adaptive Hoeffding's inequality and sequential statistical testing with early stopping rules to enhance efficiency in identifying AEs by comparing two sets of generated images---one from the original input text and the other from perturbed text characters.

\section{Enhancement of Probabilistic Robustness}
\label{sec_pr_enhance}

Despite the relatively greater focus on evaluating PR, to the best of our knowledge, limited works explicitly aimed at improving the PR of DL models are \cite{wang2021statistically,robey2022probabilistically,zhang2024prass}. \cite{wang2021statistically} demonstrates---both empirically and theoretically---that simply adding a randomly perturbed inputs significantly enhances PR and leads to better generalisation (i.e., the robustness improvement in training set can be better generalised to the test set). Both \cite{robey2022probabilistically,zhang2024prass} design loss functions using common risk measures---$\rho$-essential supremum, Conditional Value-at-Risk, and Entropic Value-at-Risk---then train models to minimize them on the training dataset.

\begin{figure}[t!]
\centering
\includegraphics[width=1\linewidth]{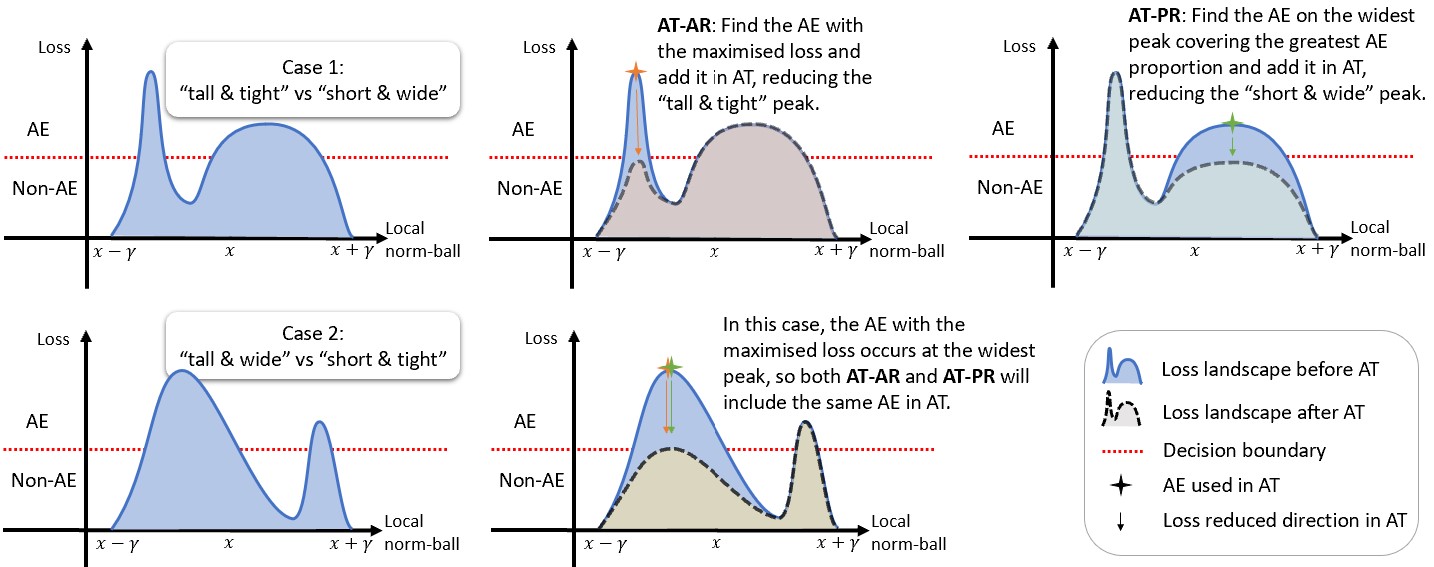}
\caption{Two cases of local loss landscape and AEs identified by AT-AR and AT-PR.}
\label{fig_idea_demo}
\end{figure}

However, these three approaches provide \textit{quite limited improvement in AR}. We believe that, akin to the well-established AT framework for AR, a dedicated AT framework for PR is necessary, which remains absent in the literature. Developing such an AT for PR scheme would \textit{enable significant joint improvement of both PR and AR}. AT \cite{goodfellow2014explaining,madry2018towards,ijcai2021p591} is a widely used empirical approach for improving robustness, typically formulated as a ``min-max'' optimisation.
The ``inner max'' often empirically solved by adversarial attacks, e.g., PGD \cite{madry2018towards} has become a critical benchmark and is widely regarded as the standard method for AT in practice \cite{ijcai2021p591}. The objective of AT for AR is to identify the AEs that maximises model loss and incorporate it into AT, cf. top-middle illustration in Fig.~\ref{fig_idea_demo}. However, to enhance local PR, the goal shifts to identifying the \textit{widest peak}---the region with the highest proportion of AEs in the loss landscape---rather than the \textit{tallest peak} (cf. top-right illustration in Fig.~\ref{fig_idea_demo}). Incorporating the AE from this region into AT can potentially reduce loss for AEs in its vicinity, thereby lowering the overall AE proportion (cf. PR interpretation in Definition~\ref{def_Prob_rob_classification}). In some cases, AT-WCR and AT-PR may select the same AE when the widest and tallest peaks coincide, as depicted in the second row of Fig.~\ref{fig_idea_demo}. The two scenarios illustrated in Fig.~\ref{fig_idea_demo} representatively cover all possible cases, though additional peaks may exist in practice.

Given the difference between the goals of improving AR and PR, we formulate a new min-max problem termed as AT-PR:
\begin{definition}[AT for PR]
\label{def_at_pr}
For a DL model \(f_\theta\), trained on a dataset \(\mathcal{S}\) consisting of pairs \((x,y)\), 
The objective of AT-PR can be formulated as:
\begin{equation}
\label{eq_at_pr_new_min_max}
\min_{\theta} \mathbb{E}_{(x,y) \sim \mathcal{S}} \left[\max_{   \|\delta\| \leq \gamma, \, \textit{PR}(x+\delta ,k)=0}  k \right],
\end{equation}
where $k$ and $\delta$ are the variables in the new inner max optimization, $\textit{PR}(\cdot,\cdot)$ is the metric defined in Eq.~\eqref{eq_probabilistic_robustness_definition}. 
\end{definition}
The intuition behind the new ``inner max'' is to find an optimal perturbation $\delta$ that produces an AE $x'=x+\delta$ that maximises the radius $k$ of a smaller norm-ball centred on this AE $x'$, such that the PR within this $k$-sized norm-ball is 0, i.e., all inputs within this new and widest norm-ball are AEs. The high non-linearity and non-convexity of DL models makes it intractable to analytically solve both the established min-max for AR and our new formulation in Eq.~\eqref{eq_at_pr_new_min_max}. Novel and \textit{efficient numerical algorithms} to solve the new framed min-max problem of AT-PR will be an important research direction.

\section{System-level Integration of Probabilistic Robustness}
\label{sec_assurance}

\begin{figure}[t!]
\centering
\includegraphics[width=0.85\linewidth]{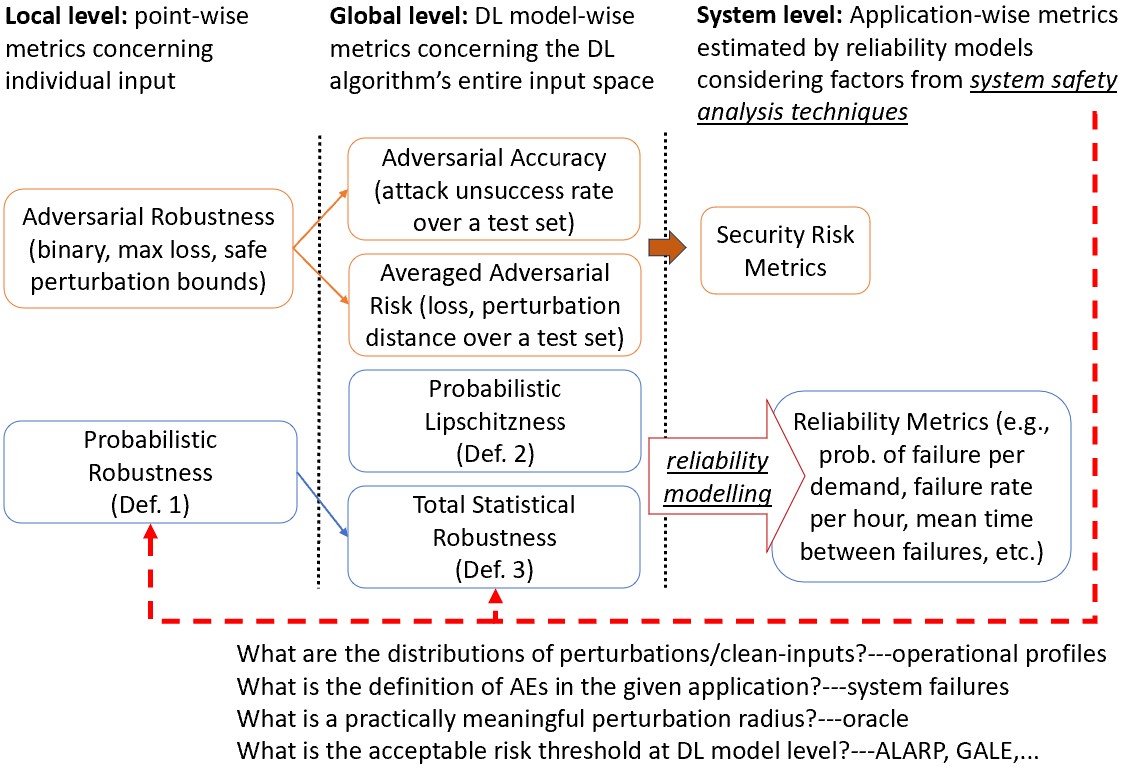}
\caption{Metrics studied ranging from local norm-ball, DL model input domain to system safety space. \textit{Safety analysis} and \textit{reliability modelling} techniques, considering factors like system architecture, redundancy design, operational environment/profiles, discrete/continuous nature, failure modes/criticality, should be applied at the system level. Then map them to application-specific estimator settings at the local and global levels of DL models, and propagate estimation results with informed confidence}
\label{fig_system_level_modelling}
\end{figure}

Safety\footnote{In system engineering, safety and reliability, though distinct, do not require separate statistical reasoning in safety-critical applications. This is because any failure in such systems can lead to unsafe outcomes. Consequently, as in \cite{zhao2019assessing}, reliability specifically refers to the probabilities/rates of failures directly impacting system safety, thereby unifying the two concepts in this context.} is a \textit{system property}, not a component property---focusing solely on ``safety at DL model levels'' without considering the full system and application context is misleading \cite{bloomfield2025ai}. A safe system accounts for redundancy, fail-safes, human oversight, and regulatory constraints, none of which are captured by evaluations at DL model levels alone. Thus, discussions of safety must extend beyond the DL model itself. As illustrated in Fig.~\ref{fig_system_level_modelling}, AR is linked to system-level security, while PR metrics can be translated into system safety through \textit{system analysis} and \textit{reliability modelling} techniques. Aligning with this perspective, several studies have explored integrating PR evidence from DL models into system-level reliability claims \cite{zhao2020safety,dong2023reliability,dong2024reachability}. However, establishing a rigorous methodology for such integration remains an open research question. Specifically, the following questions need to answered.

\paragraph{What is the most suitable reliability metric for this specific application? And what are the acceptable risk thresholds for the given metric?} Reliability in safety-critical systems is fundamentally a quantitative claim, concerned with the likelihood/rate of failures that have safety consequences. Systems can generally be categorized into continuous-time systems, which operate continuously to control a process (e.g., flight control software), and on-demand systems, which respond to discrete activation signals (e.g., the emergency shutdown system of a nuclear plant). Reliability metrics differ based on system type: failure rate (failures per unit time) is typically used for continuous-time systems, while the \textit{probability of failure per demand} (\textit{pfd}) is relevant for on-demand systems \cite{bev1993validation,zhao2020safety}. Other reliability measures may also be applicable depending on the application context, such as the probability of mission success for spacecraft launch systems, the probability of fatalities per mile and disengagement rate as studied for autonomous vehicles in \cite{kalra2016driving,zhao2019assessing,zhao2020assessing}. Safety standards such as IEC 61508 and DO-178C define domain-specific reliability metrics and establish acceptable risk thresholds based on system criticality (e.g., Safety Integrity Levels). More broadly, risk acceptability is guided by regulatory principles such as ALARP and GALE \cite{littlewood2020reliability,dong2023reliability}.

\paragraph{How can system-level reliability metrics be translated into corresponding metrics/definitions at the DL model global and local levels?} Once an appropriate system-level reliability metric is defined, reliability modelling must account for the system architecture---how the DL component interacts with traditional software and hardware, and whether ``defence-in-depth'' mechanisms, as suggested by \cite{bloomfield2025ai}, are in place. The goal is to express \textit{system reliability as a function of local/global PR-related metrics} while considering system architecture and failure dependencies. Key hyperparameters in PR-related metrics must be aligned with system-level considerations: perturbation/clean-input distributions should reflect the system’s operational profile, AE definitions should correspond to system-level failures, and perturbation radius should be derived from system specifications and testing oracles. While initial attempts have been made in \cite{dong2024reachability,zhao_assessing_2021,dong2023reliability,zhao2020safety}, these considerations must be modeled on a case-by-case basis.

\paragraph{How can PR estimates and their uncertainties be propagated from local and global levels to support system-level reliability claims with informed confidence?} Intuitively, plugging PR estimates into a reliability modelling function yields system reliability claims. However, accounting for estimation uncertainties is crucial. Fortunately, most estimators in Table \ref{tab_eva} provide confidence measures. The greater challenge lies in systematically propagating these uncertainties through the reliability model \cite{bensalem2023indeed}. This aligns with the state-of-the-art safety assurance methodology ``Assurance 2.0'' \cite{bloomfield2024confidence}, which aims for \textit{indefeasible} confidence through \textit{probabilistic assessment} and \textit{residual risk evaluation}.

\section{Challenges and Open Problems}
\label{sec_challenges}

\paragraph{Benchmarking PR estimators} 
The absence of a standardised framework for evaluating PR hinders accurate assessment and comparison of their effectiveness. Establishing rigorous benchmarking protocols is essential, requiring well-defined reference metrics, models, and datasets, along with computationally feasible gold-standard baselines. A key technical challenge lies in ensuring usability, reproducibility, and scalability across diverse models and datasets.

\paragraph{Extending PR to frontier AI tasks} PR research has predominantly focused on image classification tasks, as highlighted in Table \ref{tab_eva}, with limited exploration into other domains. Only one study has investigated text-to-image models \cite{zhang2024protip}, and another has addressed XAI \cite{Huang_2023_ICCV}. Expanding PR research to multi-modal and generative AI tasks is essential for achieving broader applicability and impact. However, unlike classification, which typically involve continuous image inputs and discrete categorical outputs, these tasks often deal with diverse modalities, e.g., images, text, and audio, as both inputs and outputs. A key open problem is the development of task-specific PR formulations that are computationally feasible while effectively capturing perturbations relevant to each modality. Furthermore, a recent work \cite{ghosh2023probabilistically} introduce PR for conformal prediction techniques.

\paragraph{Improving PR by dedicated AT} Unlike AR, which has well-established AT techniques, no dedicated AT scheme exists for PR. In this paper Sec.~\ref{sec_pr_enhance}, we have reformulated the min-max optimisation specifically for PR, establishing a new theoretical foundation for PR-focused AT. Designing AT schemes with efficient numerical algorithms to solve this new optimisation problem is challenging and forms an important future work.

\paragraph{Theoretical foundation for PR-focused AT} A critical area for future research is the theoretical analysis for PR optimised models. Understanding how PR generalises across diverse data distributions and model architectures remains an open challenge. This necessitates advancements in learning theories, particularly in areas such as Rademacher complexity and PAC-Bayesian bounds \cite{raman2022probabilistically}, to provide rigorous theoretical foundations.

\paragraph{End-to-end case studies of system-level integration} As per Sec.~\ref{sec_assurance}, PR holds significant potential for integration into system-level safety claims, but its practical applicability requires end-to-end case study demonstrations. This endeavour demands a highly interdisciplinary skill set, combining expertise in AI/DL, reliability modelling, safety assurance, and application domain knowledge.

\section{Conclusion}

This paper provides a concise yet comprehensive overview of PR in DL, covering its definitions, evaluation methods, and enhancement strategies. We introduce a reformulated AT framework tailored for PR and explore its integration into system-level safety and reliability assurance. Despite progress, key challenges remain, including benchmarking PR evaluation methods, extending PR to emerging AI tasks, and developing rigorous system-level integration methodologies. Addressing these challenges is crucial for ensuring the trustworthiness of AI in safety-critical applications.

 \bibliographystyle{splncs04}
 \bibliography{ref}

\end{document}